\documentclass[sigconf]{acmart}

\usepackage{multirow}
\usepackage[caption=false,font=normalsize,labelfont=sf,textfont=sf]{subfig}
\usepackage{textcomp}
\usepackage{stfloats}
\usepackage{amsmath,amsfonts}
\usepackage{algorithm}
\usepackage{algpseudocode}
\usepackage{array}
\usepackage{verbatim}
\usepackage{graphicx}
\usepackage{subcaption}
\usepackage{booktabs}
\usepackage{balance}
\usepackage{longtable}
\usepackage{colortbl}

\definecolor{maroon}{cmyk}{0,0.87,0.68,0.32}
\usepackage{tikz}
\newcommand*\circled[1]{\tikz[baseline=(char.base)]{
            \node[shape=circle,draw,inner sep=0.7pt,text=white,fill=darkgray] (char) {#1};}}

\AtBeginDocument{%
  \providecommand\BibTeX{{%
    \normalfont B\kern-0.5em{\scshape i\kern-0.25em b}\kern-0.8em\TeX}}}

\copyrightyear{2024}
\acmYear{2024}
\setcopyright{acmlicensed}\acmConference[MM'24]{Proceedings of the 32nd ACM International Conference on Multimedia}{October 28-November 1,2024}{Melbourne, VIC, Australia}
\acmBooktitle{Proceedings of the 32nd ACM International Conference on Multimedia (MM'24), October 28-November 1,2024, Melbourne, VIC, Australia}
\acmDOI{10.1145/3664647.3681613}
\acmISBN{979-8-4007-0686-8/24/10}

\begin{document}

\title[{Fractional Correspondence Framework in Detection Transformer}]{Fractional Correspondence Framework in Detection Transformer}

\author{Masoumeh Zareapoor}
\affiliation{%
  \institution{Shanghai Jiao Tong University }
 \streetaddress{}
  \city{Shanghai}
    \country{China}}
\email{mzarea222@gmail.com}    
 
\author{Pourya Shamsolmoali }  
\affiliation{%
  \institution{East China Normal University}
 \streetaddress{}
  \city{Shanghai}
  \country{China}}
\email{pshams55@gmail.com}

\author{Huiyu Zhou}
\affiliation{%
  \institution{University of Leicester}
 \streetaddress{}
  \city{Leicester}
 \country{United Kingdom}}
\email{hz143@leicester.ac.uk}   

\author{Yue Lu}  
\affiliation{%
  \institution{East China Normal University}
 \streetaddress{}
  \city{Shanghai}
 \country{China}}
\email{ylu@cs.ecnu.edu.cn}   
 
\author{Salvador Garc\'ia}
\affiliation{%
 \institution{University of Granada}
\streetaddress{}
 \city{Granada}
 \country{Spain}}
 \email{salvagl@decsai.ugr.es}


\renewcommand{\shortauthors}{Masoumeh Zareapoor, Pourya Shamsolmoali, Huiyu Zhou, Yue Lu, \& Salvador García}


\begin{abstract}
The Detection Transformer (DETR), by incorporating the Hungarian algorithm, has significantly simplified the matching process in object detection tasks. This algorithm facilitates optimal one-to-one matching of predicted bounding boxes to ground-truth annotations during training. While effective, this strict matching process does not inherently account for the varying densities and distributions of objects, leading to suboptimal correspondences such as failing to handle multiple detections of the same object or missing small objects. To address this, we propose the Regularized Transport Plan (RTP). RTP introduces a flexible matching strategy that captures the cost of aligning predictions with ground truths to find the most accurate correspondences between these sets. By utilizing the differentiable Sinkhorn algorithm, RTP allows for soft, fractional matching rather than strict one-to-one assignments. This approach enhances the model's capability to manage varying object densities and distributions effectively. Our extensive evaluations on the MS-COCO and VOC benchmarks demonstrate the effectiveness of our approach. RTP-DETR, surpassing the performance of the Deform-DETR and the recently introduced DINO-DETR, achieving absolute gains in mAP of {\bf{+3.8\%}} and {\bf{+1.7\%}}, respectively. 
\end{abstract}



\maketitle


\begin{figure}[t]
    \centering
    \includegraphics[width=0.77\linewidth]{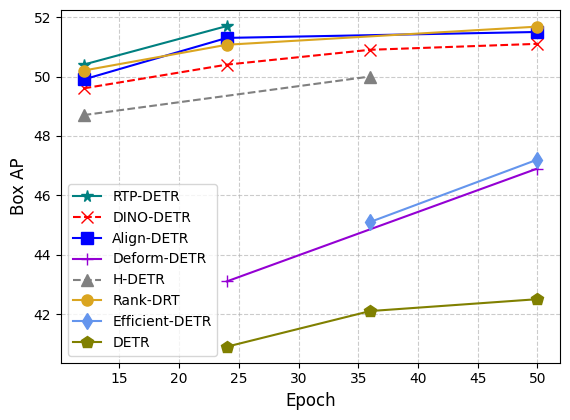}
    \caption{Learning curves ($\text{AP}$) for RTP-DETR and other DETR variants using a ResNet-50 backbone across different training durations: 12 epochs (short) and 50 epochs (long). Even with fewer epochs, RTP-DETR reaches a higher ($\text{AP}$) compared to other models, indicating its faster convergence and overall superior performance throughout the training process. DETR shows the slowest growth and lowest overall performance, converging at $\sim$42 AP by epoch 50. }
    \label{fig1}
\end{figure}

\section{Introduction}
Object detection aims to identify and localize objects within images across various categories. The advent of deep learning has significantly enhanced object detection, enabling models to achieve high accuracy and robustness across complex settings \cite{zhao2019object}. Central to the efficacy of these models is the matching process- how predictions are accurately aligned with ground-truth objects. Accurately pairing each predicted object with its ground truth is a crucial and challenging task \cite{carion2020end}. 
Conventional matching strategies in object detection, such as those used by two-stage detectors (e.g., Faster R-CNN) \cite{ren2015faster} and one-stage detectors (e.g., YOLO, SSD) \cite{redmon2016you, liu2016ssd}, rely on predefined anchor boxes and intricate overlap metrics (e.g., IoU) for aligning predictions with ground-truths \cite{detector2022fcos, zhao2019object}. Despite their widespread use, these heuristic-based methods limit the model's ability to learn optimal matching from data directly \cite{liu2023zero, hou2024salience, carion2020end}, and add complexity due to manual anchor and threshold adjustments. The Detection Transformer (DETR) \cite{carion2020end} emerges as a promising solution, introducing an end-to-end framework that simplify the object detection pipeline by using the Hungarian matching algorithm \cite{kuhn1955hungarian}. \par \noindent This algorithm provides unique one-to-one correspondences between predicted and ground-truth objects, optimizing the matching cost under the assumption of equal set sizes. In cases, where the two sets do not have the same size, significant preprocessing is required to construct a square cost matrix \cite{li2022dn, zhang2022robust, zhao2024ms}. 
DETR addresses this challenge by generating a fixed number of bounding box predictions per image, with excess predictions classified as "no object." \; However, this way can not effectively handle densely packed or significantly small objects, as the algorithm's cost function does not fully account for the global spatial \; and class distributions of objects \cite{zhao2024ms, liu2023detection, cai2023align, zhu2020deformable}.  Additionally, DETR requires long training to converge \cite{zhang2022accelerating,li2022dn,gao2021fast}, as seen in Figure \ref{fig1}. Several models addressed these issues by proposing alternatives to the Hungarian matching algorithm.  For example, Zhao et al. \cite{zhao2024ms} argue that the one-to-one assignment is not effective and attempt to combine one-to-one and one-to-many strategies to enhance the DETR performance. 

Hou et al. \cite{hou2024salience} improve the matching technique by introducing a salience score to evaluate the relationship between detected and actual objects.  Rank-DETR \cite{pu2024rank} tackled the mismatch between confidence scores and the localization accuracy of predicted bounding boxes by prioritizing predictions with more accurate localization, thereby improving the quality of matches. Despite their remarkable performance, these DETR-like detectors rely on the Hungarian algorithm for one-to-one correspondences between detected objects and ground truths, and integrate additional strategies to handle one-to-many matching scenarios. \cite{De_Plaen_2023_CVPR, ge2021ota, lin2022learning, wei2023guide} replaced the Hungarian algorithm with optimal transport (OT) to compute soft, global assignments (matching) efficiently via the Sinkhorn algorithm, avoiding the cubic complexity of the Hungarian algorithm. 



We also build upon these methods by looking for a reliable way to match sets of predictions with ground truths of potentially diverse sizes. We preserve the core property of the Hungarian algorithm—minimizing the assignment cost—while extending its capabilities to address fractional assignments and distribution discrepancies efficiently. Our matching technique is based on OT theory \cite{peyre2019computational}, optimizing the transport plan to minimize the cost across the entire distribution rather than focusing on individual matches. To be more specific, we use KL-divergence on predictions and ground truths separately to better model task-specific, non-uniform distributions. This modification ensures that the transport plan is more robust to noisy or complex assignments, resulting in improved matching accuracy.  Additionally, leveraging the Sinkhorn algorithm allows our method to maintain computational efficiency during inference.
Central to our model is the computation of a transport plan, $\Gamma$, where each element $\Gamma_{i j}$ represents the weight (degrees of matching) between predicted object $i$ and ground-truth $j$. Unlike the Hungarian algorithm, which treats matching as a binary decision (match 1, otherwsie 0), the transport plan assigns weights to each pair of prediction-ground truth, capturing the degree of match. By leveraging these weighted assessments, our approach enhances object detection, particularly for small or overlapping objects that may be missed under strict one-to-one matching scheme.  \\


{\bf{In summary:}} \circled{1} We demonstrate that DETR with the Hungarian algorithm suffers from slow convergence and struggles with accurate matching in complex senses. \circled{2} We propose a regularized transport plan to find the best alignment between predictions and ground-truths, showing how regularization can improve the convergence (see Figure \ref{fig3}). \circled{3} Our experimental results show that our model outperforms existing DETR-based methods, including Deform-DETR \cite{zhu2020deformable}, DN-DETR \cite{li2022dn}, Rank-DETR \cite{pu2024rank}, and other training-efficient variants DINO-DETR \cite{zhang2022dino} and Stable-DETR \cite{liu2023detection}.


\section{Matching Flexibility }\label{sec3}
\begin{figure*}[t]  
    \centering
    \includegraphics[width=0.73\linewidth]{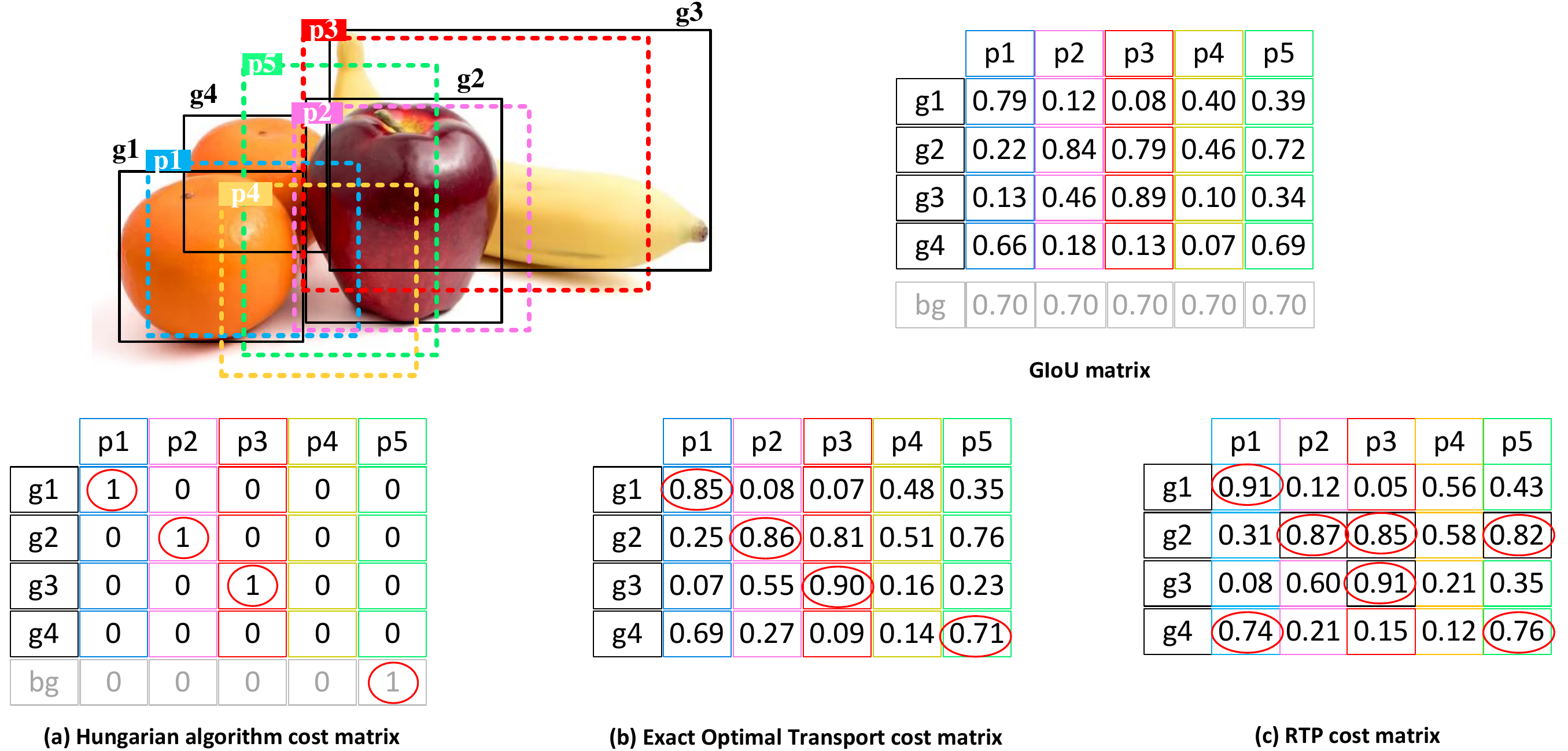}
    \caption{Cost matrix showing assignment weights (\(\Gamma_{ij}\)) between predictions \(\color{blue}{p_1}, \color{red!30}{p_2}, \color{red}{p_3}, \color{yellow}{p_4}, \color{green}{p_5}\) (dashed colored lines) and ground-truth objects \(g_1, g_2, g_3, g_4\) (solid black lines) for different matching strategies: the hungarian algorithm, exact optimal transport (without entropy regularization), and regularized transport plan (RTP). The Hungarian algorithm enforces strict one-to-one assignments, treating excess predictions as background (bg). That means, for each ground-truth object, exactly one prediction is assigned, and all other predictions are ignored (0 in the matrix), and the excess predictions are treated as background (see \(\color{green}{p_5}\rightarrow \color{gray}{bg}\)).  OT without entropy regularization tends to focus heavily on high-cost matches, thus providing hard (one-to-one) mapping; \(g_2\) is only matched to {\color{red!30}{$p_2$}} while it overlapped with {\color{red}{$p_3$}} as well. RTP uses entropy regularization to smooth assignments; each ground truth \(g_i\) connects to multiple \(p_i\) with weights distributed more evenly across the predictions; \(g_4\) has significant matches with {\color{blue}{\(p_1\)}} (0.75) and {\color{green}{\(p_5\)}} (0.76). Unlike the Hungarian algorithm, which forces all predictions to be assigned (including to the background), OT can leave predictions unassigned if they lack a sufficiently low-cost match to any ground truth.   } 
           \label{fig2}
\end{figure*}

In any matching process, it is essential to establish a matching cost that quantifies the degree of alignment between two sets. Figure \ref{fig2} compares various matching strategies—Hungarian algorithm, optimal transport without entropy regularization, and regularized transport plan (RTP)—on an image containing multiple densely packed objects. The matching cost for each pair of prediction and ground-truth bounding boxes is calculated using the Generalized Intersection over Union (GIoU) following approaches \cite{carion2020end, pu2024rank, jia2023detrs}. Since GIoU ranges from [-1, 1], we rescale it to [0, 1] for simplicity. 
(a) Hungarian cost matrix (top left): The matrix represents strict one-to-one assignments between ground-truth (\(g_1, g_2, g_3, g_4\)) and predictions (\(p_1, p_2,\cdots, p_6\)) objects. Each ground truth pairs with exactly one prediction through binary assignments (1 for a match and 0 otherwise). For example, \(g_1\) is matched to \(p_1\) while \(g_2\) is matched to (\(p_6\)), ignoring other potential matches. When a prediction does not align with a ground-truth object, it either treats it as a false negative or assigns it to the background. This precise assignment works best for datasets without overlapping or ambiguous predictions and fails to handle cases where a single prediction overlaps multiple ground truths or vice versa. Indeed, DETR often generates more predictions than actual ground-truth objects; thus, background cost acts as a balancing factor, ensuring that excess predictions are not falsely matched to ground-truth objects \cite{carion2020end, zhu2020deformable, ge2021ota, li2022dn}. 

(b) Optimal transport without entropy regularization (top middle): The transport plan tends to be sparse and deterministic, meaning each ground-truth object is mapped to a prediction, resulting in a 'hard' assignment. This behavior resembles the Hungarian algorithm, producing sharp and unambiguous couplings between source and target points. As highlighted in \cite{chuang2023infoot, chang2022unified, liu2022sparsity}, OT without regularization $\epsilon=0$, leading to high-cost assignments that may fail to account for probabilistic or fractional matches, making it less robust to noise or distributional variations. 
(c) When entropy regularization is added (RTP), the model allows soft (fractional) assignments, and the predictions can contribute to multiple ground-truth objects (probabilistic matching). The matching cost decreases because the model can distribute mass across multiple assignments (the assignments are more flexible), meaning it no longer has binary values (0 or 1). These results are optimal for achieving a balance between precision and recall, aligning with findings from \cite{De_Plaen_2023_CVPR, chuang2023infoot, wei2023guide, lin2022learning, paty2020regularized}. This analysis underscores the advantages of regularized OT over the Hungarian algorithm, particularly in handling discrepancies between the number of predictions and ground truths.

\section{Related Work}
\subsection{DETR with different matching frameworks}
DETR \cite{carion2020end} revolutionized object detection by presenting it as a set prediction problem, using one-to-one matching supervised by the Hungarian algorithm for end-to-end training. Several subsequent works have proposed to address the slow convergence of DETR from different perspectives. \cite{sun2021sparse} stated that the cross-attention mechanism in the decoder is the bottleneck to training efficacy, and suggested an encoder-only architecture as a solution. Gao et al.\cite{gao2021fast} aimed to streamline the cross-attention process by integrating a Gaussian before regulating attention within the model. 
Another direction to enhance DETR is to refine its matching strategy, which is more relevant to our work. This focus stems from the critical role of matching in object detection \cite{hou2024salience, detector2022fcos}, where accurately pairing each predicted object with its ground-truth during training (as illustrated in Figure \ref{fig2}).  Deform-DETR \cite{zhu2020deformable} enhances the DETR's efficiency and performance, particularly when dealing with small objects or objects that require finer spatial resolution. This model uses Hungarian matching via a new optimization technique to find the best one-to-one correspondence between predicted and ground-truth objects during training. \cite{hou2024salience} proposed Salient-DETR, a salient score between detected and actual objects to ensure that only the most relevant objects are matched. 
Stable-DETR \cite{detector2022fcos} reveals that DETR's slow convergence and performance issues stem from an unstable matching problem. It addresses this by introducing a stable matching technique that prioritizes positional metrics over semantic scores. DN-DETR \cite{li2022dn} improves accuracy by using parallel decoders with shared weights to process multiple noisy queries derived from ground-truth objects. Group DETR \cite{chen2022group} enhances DETR by adding multiple decoder groups, each handling specific subsets of object queries. Both DN-DETR and Group DETR use one-to-one matching strategy for every group of object queries to ensure accurate alignments. Meanwhile, DINO \cite{zhang2022dino} advances this by incorporating dynamic anchors and denoising training, achieving the first state-of-the-art performance on the COCO benchmark among DETR variants. Different from these approaches, our model seeks optimal alignment by computing transport plans between pairs of prediction and ground-truth objects. 

\subsection{Optimal Transport Alignment}
Originally, optimal transport (OT) \cite{peyre2019computational} tackles the problem of finding the most cost-effective way to align two sets of points (or distributions). It looks for an optimal coupling (transport plan) between distributions $\mu$ and $\nu$, representing it as a joint probability distribution. In other words, if we define $U(\mu, \nu)$ as the space of probability distributions over $R^d$ with marginals $\mu$ and $\nu$, the optimal transport is the coupling $\Gamma \in U(\mu, \nu)$, which minimizes the following quantity: $\min \langle\Gamma, C_{i j}\rangle$, where $C_{i j}$ is the cost of moving \(i\) (from $\mu$) to \(j\) (from $\nu$), respectively. The use of OT has gained popularity in generative modeling \cite{shamsolmoali2022gen}, adversarial training \cite{wong2019wasserstein}, and domain adaptation \cite{chang2022unified}, and many other disciplines \cite{chuang2023infoot, shamsolmoali2024setformer, shamsolmoali2024hybrid}, since the introduction of Sinkhorn’s algorithm \cite{cuturi2013sinkhorn}. OT and the Hungarian algorithm, while both aimed at solving the assignment problem by finding the best matches between elements of two sets, differ significantly in their methodological execution. {{Replacing the Hungarian algorithm with OT for finding alignments between distributions has been explored in the literature \cite{De_Plaen_2023_CVPR,ge2021ota,clark2022unified,liu2022sparsity,lin2022learning, wei2023guide, xie2022solving}}}. These studies demonstrate OT’s effectiveness in scenarios where exact matching fails to capture the complexity of relationships, such as dense object detection or multi-modal distributions. 


\section{Method}
We first revisit important details on the Hungarian algorithm and OT, which will be useful to describe our proposed model. 

\subsection{Notations}
For each image, we have a set of ground-truth objects \(\mathcal{O}^*=\{o_1^*, o_2^*, \\ \ldots, o_N^*\}\), and a set of predicted objects $\hat{\mathcal{O}}=\left\{\hat{o}_1, \hat{o}_2, \ldots, \hat{o}_M\right\}$, where $N$ is the number of ground-truth objects and $M$ is the number of predicted objects ($M \geq N$). Each ground-truth $o_i^*$ and predictions $\hat{o}_j$ is represented by a combination of class label and bounding box coordinates, denoted as $o_i^*=[c_i^*, b_i^*]$ and $\hat{o}_j=[\hat{c}_j, \hat{b}_j]$, respectively. Throughout the paper, vectors are denoted by lowercase letters, and matrices by uppercase. $1_N$ is $N$-dimensional vectors of ones, and $1_{M\times N}$ denotes an $M\times N$ matrix, each element of which is 1. The probability simplices  $\Delta^N$ and  $\Delta^M$, defined as  $\Delta^N:=\{u\in \mathbb{R}^{N}: \sum_i u_i=1\}$ and $\Delta^M:=\{v\in \mathbb{R}^{M}: \sum_j v_j=1\}$, represent the sets of all possible weights for discrete measures across $N$ and $M$.

\subsection{Hungarian matching algorithm}
DETR uses the Hungarian algorithm \cite{kuhn1955hungarian} to establish an one-to-one pairing between predicted and actual objects by minimizing matching costs. The objective is to find an optimal permutation $\sigma$ of the $M$ predictions that minimizes the total matching cost:
\begin{equation}
\sigma^*=\underset{\sigma}{\operatorname{argmin}} \sum_{i=1}^N C\left(\mathcal{O}_i^*, \hat{\mathcal{O}}_{\sigma(i)}\right),
\label{eq1a}
\end{equation}
where $C(\mathcal{O}^*, \hat{\mathcal{O}})$ is the cost of matching the \(i\)-th ground-truth object to a prediction indexed by $\sigma(i)$. ${C}$  is a weighted sum of a classification loss and a localization loss (bounding boxes), defined as
\begin{equation}
{C}_{i j}=-\log(p_{j}(c_i))+ \lambda_{bbox} {L}_{bbox}(b_i^*, \hat b_j)+\lambda_{GIoU}(1-\text{GIoU} (b_i^*, \hat b_j)),
\label{eq1}
\end{equation}
in which $p_{j}(c_i)$ represents the probability that the \(j\)-th prediction correctly classifies the \(i\)-th ground truth. ${L}_{bbox}$ and $\text{GIoU}$ measures the localization error between predicted and actual bounding boxes, with $\lambda$ parameters tuning the importance of each component. While the Hungarian algorithm ensures unique pairings to minimize the matching cost, it cannot always align with the complex realities of object detection, where multiple predictions correspond to a single object due to overlaps or visual ambiguities \cite{zhao2024ms, hou2024salience, chen2022group}. Moreover, this matching approach does not consider the overall distribution of predictions or ground truths, which means it evaluates matches individually, ignoring the broader context of how all predictions and ground truths should ideally be distributed or matched \cite{hou2024salience, pu2024rank}.

\subsection{Optimal Transport (OT)}
In the quest to enhance DETR performance, we use OT \cite{peyre2019computational}, a mathematical formulation that has been widely applied in various alignment problems. OT aims to minimize the cost of transporting `mass' from one distribution to another. In the context of object detection, OT can align a set of predicted objects to a set of ground-truth objects. Each `mass' corresponds to an object, whether a prediction or a ground truth. Consider the ground-truth objects $\{o_i^*\}_{i=1}^N$ and the predicted objects $\{\hat{o}_j\}_{j=1}^M$, with their respective distribution weights $\mu\in \Delta^N$ and $\nu\in \Delta^M$. Here, $\mu$ and $\nu$ denote the importance or confidence of each ground truth and prediction, located at $o_i^*$ and $\hat{o}_j$, respectively. We then use a transportation cost $C \in \mathbb{R}^{M\times N}$ to compute a plan $\Gamma$ that minimizes the total cost of matching each prediction $j$ to a ground truth $i$
\begin{equation}
\begin{aligned}  
 \min_{\Gamma\in {U}(\mu, \nu)} & \langle \Gamma, \mathbf{C}\rangle_F,  \\
\textbf{where} \quad  U(\mu, \nu)=\{\Gamma \in \mathbb{R}_{+}^{M\times N} & :  \Gamma \mathbf{1}_N=\mu, \quad \Gamma^T \mathbf{1}_M=\nu\}, 
\label{eq3}
\end{aligned}
\end{equation}
$\langle\Gamma, \mathbf{C}\rangle_F$ is the Frobenius dot product between the transport plan and cost matrix. The cost $\mathbf{C}\geq0$ represents how well (or poorly) a predicted object matches with a ground truth. The polytope $U(\mu, \nu)$ is a set of transportation plans of dimension $M\times N$ to match predictions with ground truths, ensuring an optimal alignment under specified constraints. Each element $C_{ij}$ in the cost matrix is computed using the Wasserstein distance (detailed in \cite{paty2020regularized,chuang2023infoot}). However, the Hungarian algorithm does not account for the distributions $\mu$ and $\nu$ of ground truths and predictions, treating all matches equally \cite{cai2023align, pu2024rank, De_Plaen_2023_CVPR}. It enforces a strict one-to-one matching, leading to situations where some predictions may not match any ground truth, thus deemed to match the background. OT, on the other hand, lets several predictions use the distribution of each ground-truth object and vice versa, making the matching process more flexible. 
\begin{figure}
    \centering
   \includegraphics[width=0.88\linewidth]{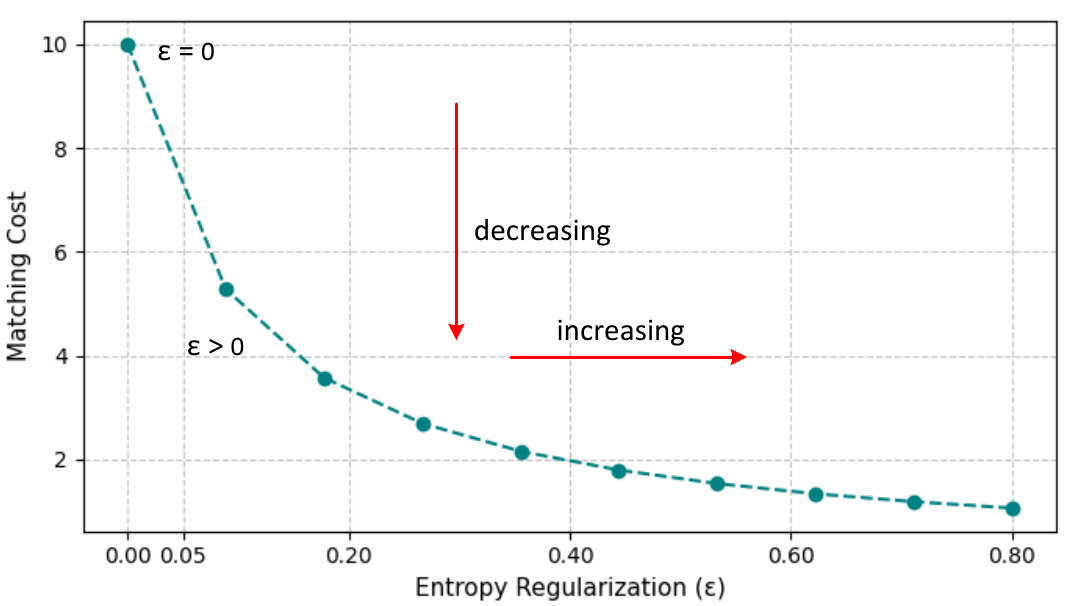}
    \caption{Effect of using regularization $H(\Gamma)$. The x-axis represents the matches $\Gamma$ (transport plan), which pairs predicted objects with ground-truth objects. The y-axis is the matching cost, with lower values denoting more cost-effective pairings. The regularized transport plan (RTP) (with $\epsilon \neq 0$) shows reduced matching costs, indicating the benefits of the regularization term $H(\Gamma)$ in achieving a smooth distribution of matches. Interestingly, at $\epsilon=0$, the transport plan is sharp and this rigidity forces higher-cost assignments because the algorithm cannot "distribute" assignments to reduce cost.}
    \label{fig3}
\end{figure}

\subsection{Regularized Transport Plan (RTP)}
It is well-known that OT generates fully dense transportation plans, meaning that every prediction is (fractionally) matched with all ground truths \cite{liu2022sparsity, chuang2023infoot, chang2022unified}. This matching flexibility, however, introduces the problem of over-splitting. This occurs when the mass is spread too tiny across multiple predictions or ground truths, hence reducing the accuracy and reliability of the matching process. Indeed, OT alone cannot provide an effective alignment \cite{sebbouh2023structured}. We incorporate a Kullback-Leibler (KL) divergence to relax the strict conservation of marginal constraints in OT by implementing soft penalties \cite{chang2022unified}. This adjustment helps to control the distribution of probabilities, preventing over-splitting and ensuring that the matching process remains both accurate and fair. Given distributions $\mu \in\mathbb{R}^N$ for ground truths and $\nu \in \mathbb{R}^M$ for predictions, along with a cost matrix $C$ that $C_{ij}$ reflects the cost of matching predictions $j$ to ground-truth $i$, our goal is to find a transportation plan $\Gamma \in \mathbb{R}^{M \times N}$ that minimizes the total cost while ensuring that the resulting distribution of predictions and ground truths (after transportation) closely aligns with their original distributions:
\begin{equation}
\min _{\Gamma \geq 0}\langle C, \Gamma\rangle + \kappa_1 D_{KL} (\Gamma \mathbf{1}_N \| \nu) + \kappa_2 D_{KL} (\Gamma^T \mathbf{1}_M \| \mu),
\label{kl}
\end{equation}
where $D_{\text{KL}}$ is KL divergence. $\langle C, \Gamma\rangle=\sum_{i=1}^N \sum_{j=1}^M C_{i j} \Gamma_{i j}$ is the total transport cost, and $\kappa_1$, $\kappa_2$ are regularization parameters that balance the KL divergence terms. These terms align the transported mass distributions ($\Gamma^T 1_M$ and $\Gamma1_N$) with the original distributions ($\nu$ and $\mu$), allowing the model to address discrepancies in the total distribution between predictions and ground-truths efficiently. The KL divergence measures the difference between the transported distribution ($\mathbf{1}_M^T, 1_N$) and the original distribution ($\mu, \nu$), as follow
\begin{equation}
\operatorname{KL}(\Gamma^T\mathbf{1}_M\|\mu)=\sum_{i=1}^M (\Gamma^T\mathbf{1}_M)_i\log(\frac{(\Gamma\mathbf{1}_M)_i}{\mu_i}),
\end{equation}
for predictions, ensuring alignment with $\nu$, and 
\begin{equation}
\operatorname{KL}(\Gamma\mathbf{1}_N\|\nu)=\sum_{j=1}^N(\Gamma\mathbf{1}_N)_j\log(\frac{(\Gamma^T\mathbf{1}_N)_j}{\nu_j}),
\end{equation}
for ground truths, ensuring alignment with $\mu$. By embedding KL divergence, the finalized transport plan adheres closely to the original distributions of both predictions and ground-truths. This ensures that the matching process respects the inherent probabilistic nature of detections, accommodating scenarios where the number of predictions exceeds or falls short of the number of ground-truth.

\subsubsection{Mass Constraints with Entropy Regularization}
We now consider an entropic regularization of transport plan $\Gamma$ that controls the smoothness of the coupling and is particularly useful for reducing the computational complexity \cite{cuturi2013sinkhorn, chang2022unified, sebbouh2023structured}. Sinkhorn's algorithm is used to compute regularized OT. This algorithm is known for being fast and being able to differentiate based on inputs. Furthermore, the resulting transport plan is easier to interpret because it provides a probabilistic view of the relationships between predicted and actual objects. The regularized version of Eq. (\ref{kl}) is
\begin{equation}
\begin{aligned}
&\min _{\Gamma}\sum_{i=1}^N\sum_{j=1}^M C_{i j}\Gamma_{i j}+\kappa_1 \cdot \mathrm{KL}\left(T\mathbf{1}_N, \nu\right) \\&+\kappa_2 \cdot \mathrm{KL}\left(T^T \mathbf{1}_M, \mu\right) -\epsilon H(\Gamma),
\label{eq7}
\end{aligned}
\end{equation}
where $H(\Gamma)=-\sum_{ij} \Gamma_{i j} \log(\Gamma_{i j})$ is the entropy of the transport plan $\Gamma$ and $\epsilon>0$ controls the smoothness of the transport plan. Entropy regularization plays a critical role in ensuring numerical stability and enabling efficient computation through Sinkhorn’s algorithm. However, reducing to ($\epsilon=0$) eliminates these benefits, as the solution approaches the exact OT formulation. In this case, the transport plan becomes deterministic, resulting in one-to-one mappings, as discussed in \cite{paty2020regularized} (Proposition 2) and \cite{chuang2023infoot} (Discrete vs. Continuous). This behavior, illustrated in Figure \ref{fig3}, highlights the trade-off between the precision of OT and the scalability and flexibility of entropy-regularized OT. Adding entropy ($\epsilon>0$) allows the transport plan to become smoother thus, we can assign predictions fractionally to multiple ground-truth objects (and vice versa).



\begin{algorithm}[t]
\caption{Regularized Transport Plan (RTP) Matching}\label{alg1}
\begin{algorithmic}[1]
\Require $C$: cost matrix ($M\times N$), $\mu$: distribution of ground-truths ($N$-vector), $\nu$: distribution of predictions ($M$-vector), $\kappa$: regularization parameter for KL divergence, $\epsilon$: regularization parameter for entropy
\Ensure $\Gamma$: Optimal Assignment ($M\times N$ matrix)
\State $K[i, j] \gets \exp(-C[i, j] / \epsilon) $ \Comment{\textcolor{orange}{Initialize Gram matrix}}
\State $u \gets \textbf{1}_N, \; v \gets \textbf{1}_M$ \Comment{\textcolor{orange}{Initialize scaling vectors}}
\While{not converged} \Comment{\textcolor{orange}{Sinkhorn algorithm }}
    \State $u[i] \gets \mu[i] / (K v)[i] $ \Comment{\textcolor{orange}{Update scaling vector $u$ $\in$ $\mu$}}
    \State $v[j] \gets \nu[j] / (K^\top u)[j] $ \Comment{\textcolor{orange}{Update scaling vector $v\in \nu$ }}
\EndWhile
\State $\Gamma[i, j] \gets u[i] \cdot K[i, j] \cdot v[j] $ \Comment{\textcolor{orange}{Compute transport plan}}
\For{$i \leftarrow 1$ to $N$} \Comment{\textcolor{orange}{Adjust $\Gamma$ for KL divergence w.r.t. $\mu$}}
    \State $\text{rowSum} \leftarrow \sum_{j}\Gamma[i, j]$ 
    \If{$\text{rowSum} \neq \mu[i]$}
        \State $\Gamma[i, :]$ to make the row sums align with $\mu[i]$.
    \EndIf
\EndFor
\For{$j \leftarrow 1$ to $M$} \Comment{\textcolor{orange}{Adjust $\Gamma$ for KL divergence w.r.t. $\nu$}}
    \State $\text{colSum} \leftarrow \sum_{i}\Gamma[i, j]$
    \If{$\text{colSum} \neq \nu[j]$}
        \State $\Gamma[:, j]$ to make the column sums align with $\nu[j]$.
    \EndIf
\EndFor
\State \Return $\Gamma$
\end{algorithmic}

\end{algorithm}

\section{Experimental Results}
This section begins with training details (Sec. \ref{train}). Sec. \ref{sub2}, Sec. \ref{sub3}, and Sec. \ref{sub4} then present comparisons with state-of-the-art methods and convergence analysis. Finally, Sec. \ref{sub5} brings ablation.

\subsection{Training settings} \label{train}
We evaluated our matching strategy, with a time complexity of \(\mathcal{O}(T\cdot(N\cdot M))\), where \(T\) is the number of iterations required for convergence, \(N\) is the number of ground-truth objects, and \(M\) is the number of predictions. The first dataset, COCO object detection \cite{lin2014microsoft}, includes 118,287 training images and 5,000 validation images. The second dataset, PASCAL-VOC object detection, is fine-tuned on VOC trainval07+12 with approximately 16.5k images and evaluated on the test2007 set. We compared our model (RTP-DETR) against several state-of-the-art DETR variants, including DN-DETR \cite{li2022dn}, Salience-DETR \cite{hou2024salience}, Rank-DETR \cite{pu2024rank}, DINO-DETR \cite{zhang2022dino}, and Group-DETR \cite{chen2022group}. All models were trained with the same set of hyperparameters: $\lambda_{CrossE}=1$, $\lambda_{GIoU}=2$ and $\lambda_l=5$. As noted in \cite{cuturi2013sinkhorn}, using a fixed entropy regularization parameter $\epsilon$ is problematic because the number of predictions, object densities, and cost variations change dynamically across images and datasets. Instead of using a fixed one $\epsilon$, we use an adaptive function that scales based on the number of samples (in our case, M is the number of predictions), as $\epsilon=\epsilon_0 /{\log M}$. This ensures that when 
\(M\) is large, $\epsilon$ decreases smoothly, preventing excessive regularization; when \(M\) is small, $\epsilon$ remains stable, avoiding overly aggressive smoothing. Moreover, to enforce marginal constraints in the transport plan, we use two complementary weights: $\kappa_2$, and $\kappa_1=1-\kappa_2$, which regulate the influence of ground-truth and prediction distributions, respectively. Indeed, this relationship ensures that the weights $\kappa_1$, $\kappa_2$  are complementary: when one $\kappa_1$  increases, the other $\kappa_2$  decreases, and vice versa, and the sum of $\kappa_1+\kappa_2=1$ remains constant.

\begin{table*}[t]
    \centering
    \begin{tabular}{l|c c| c c c c c c c}  \toprule
    Method & Backbone & \#epochs & AP $\uparrow$ &  $\text{AP}_{50}$ $\uparrow$ & $\text{AP}_{75}$ $\uparrow$ & $\text{AP}_S$ $\uparrow$ &  $\text{AP}_{M}$ $\uparrow$ & $\text{AP}_{L}$ $\uparrow$ & Avg. (\%) \\ \midrule 

Deform-DETR \cite{zhu2020deformable} & ResN50 & 50 &  46.9 & 65.6 & 51.0 & 29.6 & 50.1 & 61.6 & 50.8   \\ 

Sparse-DETR \cite{roh2021sparse} & ResN50 & 50 & 46.3 & 66.0  &  50.1  & 29.0 &   49.5  &  60.8 & 50.3  \\

Effcient-DETR \cite{yao2021efficient} & ResN50 & 36 & 45.1 &  63.1 &  49.1 &  28.3 &  48.4  & 59.0 & 48.8   \\

H-DETR \cite{jia2023detrs} & ResN50 & 36 &50.0 &68.3& 54.4 &32.9 &52.7 &65.3  & 53.9      \\ 

DN-DETR \cite{li2022dn} & ResN50 & 12 & 43.4 & 61.9 & 47.2 & 24.8 & 46.8 & 59.4 & 47.2  \\ 

Rank-DRT \cite{pu2024rank} & ResN50 & 12 & 50.2 & 67.7 & 55.0 & 34.1 & 53.6 & 64.0 & 54.1 \\ 

DINO-DETR \cite{zhang2022dino} & ResN50 &  12 & 49.7 & 66.6 & 53.5 & 32.0 & 52.3 &  63.0 & 52.7  \\ 

Salience-DETR \cite{hou2024salience} & ResN50 & 12 & 49.4 &67.1& 53.8 &32.7 &53.0 &63.1 & 53.2  \\ 

H-DETR \cite{jia2023detrs} & ResN50 &12 &48.7 &66.4& 52.9 &31.2 &51.5 &63.5  & 52.4     \\ 

\midrule

\rowcolor{maroon!15}
RTP-DETR & ResN50 &  12 & 50.4 & 67.9 & 55.2 & 34.7 & 53.8 & 64.2  &  54.4  \\ 

  \bottomrule
    \end{tabular} 
    \caption{Comparison of our approach (RTP-DETR) with top-performing DETR-based models using the ResNet50 backbone on the COCO dataset. Our model consistently surpasses or performs competitively with respect to the state-of-the-art baselines.}
    \label{tab2}
\end{table*}
\begin{table*}[t]
    \centering
    \begin{tabular}{l|lcc|ccccccc} \toprule
   Method & Matching & Backbone& \#epochs & AP $\uparrow$ & $\text{AP}_{50}$ $\uparrow$& $\text{AP}_{75}$ $\uparrow$& $\text{AP}_S$ $\uparrow$ & $\text{AP}_M$ $\uparrow$& $\text{AP}_L$ $\uparrow$ & Avg. (\%) \\ \hline
   
\color{blue} DINO-DETR \cite{zhang2022dino} &  & \color{blue}ResN50 & \color{blue}36 & \color{blue}50.9& \color{blue}69.0 & \color{blue}55.3 & \color{blue}34.6 & \color{blue}54.1 & \color{blue}64.6  & \color{blue} 54.7   \\ \hline
 
\multirow{3}{*}{DINO-DETR \cite{zhang2022dino}}&  {Baseline} & ResN50 & 12 &49.7 &66.6 &53.5 &32.0 &52.3 &63.0 & 52.7   \\
                                               & Rank \cite{pu2024rank} & ResN50 & 12 & 50.4 & 67.9 & 55.2 & 33.6 & 53.8 & 64.2 & 54.2  \\ 
                                               & Stable-DINO \cite{liu2023detection} & ResN50 & 12 & 50.3 & 67.4 & 55.0 & 32.9 & 54.0 & 65.5 & 54.2  \\ 
                                       
 \rowcolor{maroon!15}
 
\cellcolor{white} & RTP & ResN50 & 12 & 50.4 & 68.2 & 55.2 & 33.5 & 54.0 & 65.5 & 54.5       \\ \hline

\color{blue}{DINO-DETR \cite{zhang2022dino}} &  & \color{blue}Swin-L & \color{blue}36 & \color{blue}58.0 & \color{blue}77.1 & \color{blue}66.3 & \color{blue}41.3 & \color{blue}62.1 & \color{blue}73.6 &\color{blue} 63.1  \\ \hline

\multirow{3}{*}{DINO-DETR \cite{zhang2022dino}} & Baseline & Swin-L & 12 & 56.8 & 75.4 &62.3 &40.0 &60.5 &73.2 & 61.4   \\  
   & Rank \cite{pu2024rank} & Swin-L & 12 & 57.5 & 76.0 & 63.4 & 41.6 & 61.4 & 73.8 & 62.3  \\   
 & Stable-DINO \cite{liu2023detection} & Swin-L & 12 & 57.7 & 75.7 & 63.4 & 39.8 & 62.0 & 74.7 & 62.2  \\ 
 
 \rowcolor{maroon!15}
 
\cellcolor{white} & RTP & Swin-L & 12 & 57.9 & 76.1 & 63.6 & 41.5 & 62.3 & 74.9 & 62.7 \\ \bottomrule   
\end{tabular} 
    \caption{Enhancing object detection performance with DINO-DETR on the COCO val2017 dataset, using ResNet50 \cite{he2016deep} and Swin-Large \cite{liu2021swin} backbones. RTP-DETR is presented as a complementary model to existing methods and achieves consistent enhancements in performance. We also include the result of DINO-DETR trained for 36 epochs as a reference for comparison. }
    \label{dino}
\end{table*}

\vspace{-5pt}
\subsection{Comparison with DETR-based methods}\label{sub2}
Table \ref{tab2} shows the performance of RTP-DETR compared with other high-performing DETR-based methods on the COCO object detection val2017 set using ResNet-50 \cite{he2016deep}. With only 12 training epochs, RTP-DETR achieves an impressive AP of 50.4\%, which suppresses H-DETR by +2.0\% and exceeds the most recent state-of-the-art Salience-DETR and Rank-DETR, by +1.3\% and +0.5\%. Interestingly, we observe notable improvements in $\text{AP}_{75}$, demonstrating the advantage of our approach at higher thresholds. Sparse-DETR and Deform-DETR (Avg: 50.3 and 50.8) have lower average performance compared to the top performers. Additionally, Table \ref{dino} illustrates the effectiveness of our model in enhancing DINO-DETR, a training-efficient DETR variant that has received significant attention in the object detection task. The results are obtained using two different backbones, ResNet-50 \cite{he2016deep} and Swin-Large \cite{liu2021swin}. We also compare our model with other integrations such as Rank-DINO and Stable-DINO. The improvements with our model are notable: there's an increase of 1.6\% in AP performance when using the ResNet-50 backbone (49.7\% vs. 50.4\%) and 1\% with Swin-L (56.8\% vs. 59.7\%). At a higher IoU threshold ($\text{AP}_{75}$), the enhancements become even more pronounced, achieving +1.7\% improvement with ResNet-50 and +1.3\% with Swin-L. These results indicate our model's robustness and generalizability across various DETR-based models.

\subsection{Combination with DETR variants}\label{sub3}
Table \ref{tab1} summarizes the integration of our RTP with several DETR variations. As can be seen, when combined with Deform-DETR, the AP has increased by +1.8, from 46.9\% to 48.7\%. Similarly, the performance of Group-DETR has improved by +0.7, moving from 48\% to 48.7\%, while using only 12 epochs schedule. Additionally, our RTP has enhanced DINO-DETR's performance, boosting its AP by +1.4 over a 24-epoch and by an additional +1.6 AP when the training is extended to 36 epochs. It's important to note that many of these methods rely on a one-to-one matching strategy, with modifications mainly focused on optimizing the Hungarian algorithm to better accommodate one-to-many relationships between predictions and ground truths during training. Unlike these advances, RTP utilizes a different matching strategy, providing additional advantages of regularization that contribute to the performance gains observed.

\begin{figure}
    \centering
    \includegraphics[width=0.9\linewidth]{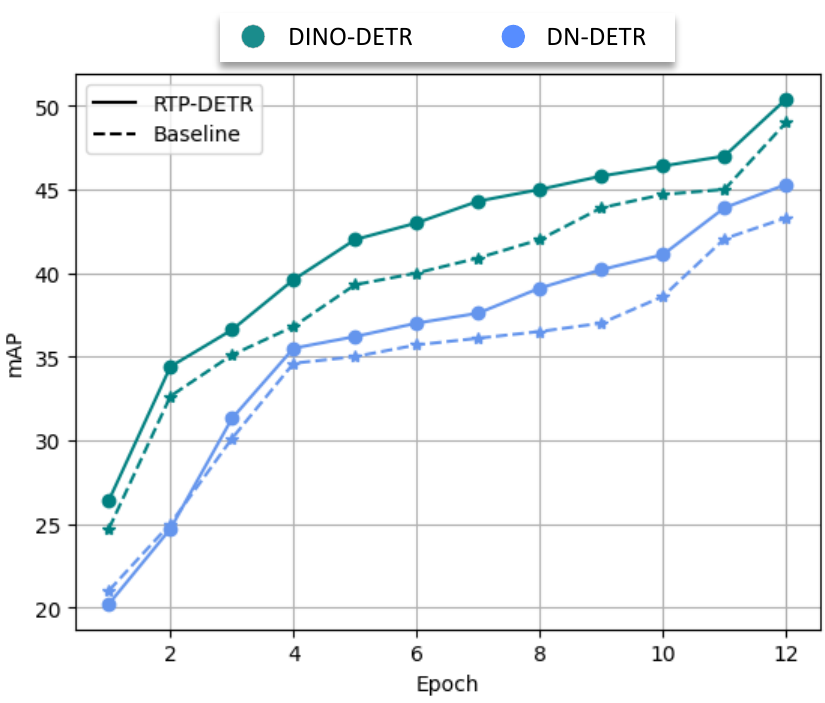}
    \caption{Convergence curves. RTP accelerates the training process for different variants of DETR. The baseline models and our RTP counterparts are shown by dotted and solid lines, respectively. The horizontal axis denotes the number of epochs, the vertical axis is the AP evaluated on COCO.}
    \label{conv}
\end{figure}

\subsection{Convergence}\label{sub4}
Adding entropy regularization into the transport plan (Eq. (\ref{eq7})), yields an efficient and seamless matching process. Figure \ref{fig3} provides a clear illustration of this improvement by showing the effectiveness and speed of the RTP matching process. Furthermore, Figure \ref{fig1} details the convergence behavior of our model, demonstrating that RTP enables significantly faster convergence compared to standard DETR. The slower convergence rates of conventional DETR can be attributed to the discrete and unstable nature of the Hungarian algorithm, particularly during the early stages of training \cite{li2022dn, zhu2020deformable, zhang2022accelerating}. We also conducted a comparative analysis of our model alongside two prominent DETR variants, DINO \cite{zhang2022dino} and DN \cite{li2022dn}, known for their effective one-to-many matching capabilities. The models use a ResNet-50 backbone and are trained over 12 epochs, as shown in Figure \ref{conv}. This comparison demonstrates that RTP not only refines the matching process but also significantly accelerates model convergence during training, thereby providing a robust solution to the inherent limitations of the Hungarian matching algorithm.

\begin{table}[]
    \centering
    \begin{tabular}{l|c c c c@{}}  \toprule
Method & Matching & \# epochs & AP $\uparrow$  & AR $\uparrow$  \\ \midrule 

Deform-DETR & -  & 50 & 46.9 & 57.3  \\ \rowcolor{maroon!15}
Deform-DETR & RTP &  50 & 48.7 \color{gray}(+1.8) & 58.6 \color{gray}(+1.3)  \\  \midrule

DINO-DETR & -  &  24  &  50.4 & 65.4     \\ \rowcolor{maroon!15}
DINO-DETR &  RTP &  24  &  51.8  \color{gray}(+1.4) & 66.3  \color{gray}(+0.9)    \\  \midrule

Group-DERTR & -  & 12 & 48.0 &   67.2                   \\ \rowcolor{maroon!15}
Group-DERTR  & RTP  & 12 & 48.7  \color{gray}(+0.7) &   67.9 \color{gray}(+0.7)  \\  \midrule

Salience-DETR & - & 12 & 49.2 & 63.5 \\  \rowcolor{maroon!15}
Salience-DETR & RTP & 12 & 49.8  \color{gray}(+0.6) & 64.0  \color{gray}(+0.5) \\    \midrule

Rank-DETR & - & 12 & 50.2 & 67.9  \\  \rowcolor{maroon!15}
Rank-DETR & RTP &  12 & 50.4  \color{gray}(+0.2) & 68.2 \color{gray}(+0.3)    \\ 
  \bottomrule
    \end{tabular} \vspace{5pt}
    \caption{Combination with other methods. RTP is a complementary approach that consistently improves performance. }
    \label{tab1}
\end{table}

\begin{table}[]
    \centering
    \begin{tabular}{c|c c c} \toprule
Method  & \# epochs & AP $\uparrow$  & AR $\uparrow$    \\  \midrule
Align-DETR \cite{cai2023align} & 12 &  50.2 & 61.7  \\   \rowcolor{maroon!15}
 + RTP & 12 & 50.6 \textcolor{gray}{(+0.4)} & 62.3 \textcolor{gray}{(+0.6)} \\ \midrule
Align-DETR \cite{cai2023align} &  24 & 51.3 & 62.4\\ \rowcolor{maroon!15}
+ RTP &  24 &  51.9 \textcolor{gray}{(+0.7)} & 62.9 \textcolor{gray}{(+0.5)}\\
\bottomrule
    \end{tabular} \vspace{5pt}
    \caption{The effect of combining RTP with Align-DETR demonstrates a significant improvement in performance.  }
    \label{abl}
\end{table}

\subsection{Ablation}\label{sub5}

\subsubsection{Computation Time}
\begin{figure}[t]
    \centering{
   \includegraphics[width=1\linewidth]{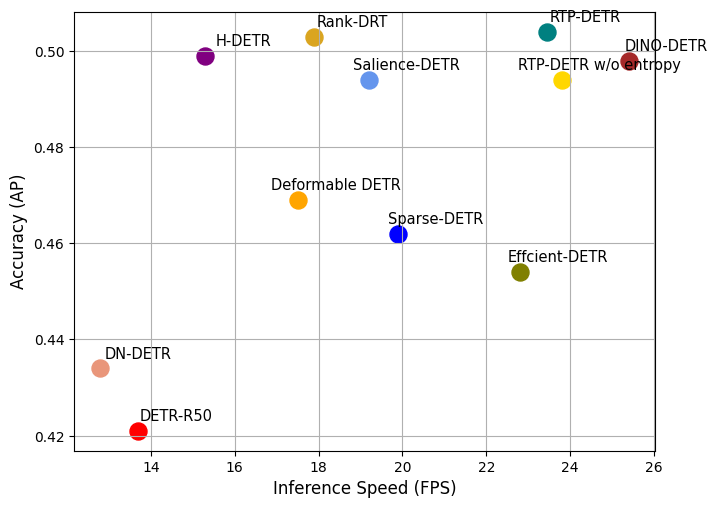}
    \caption{Inference time vs. accuracy on COCO for various DETR-based models. The x-axis represents inference speed (FPS), where higher values indicate faster models, and the y-axis represents accuracy (AP), where higher values indicate better performance. DETR and DN-DETR achieve relatively low accuracy with similar inference speeds ($\sim$12–13 FPS). Our model, RTP-DETR, achieves the highest accuracy with competitive inference speed, outperforming DINO-DETR in precision, although it exhibits slightly lower efficiency.}
    \label{time}}
\end{figure}

Figure \ref{time} compares the inference speed (FPS) and accuracy (AP) of various DETR-based models, including our RTP models with and without entropy regularization ($\epsilon>0$ and $\epsilon=0$, respectively). We observe that both RTP variants demonstrate superior performance compared to models relying on the Hungarian algorithm. This is because the Sinkhorn algorithm, which underpins optimal transport with entropy regularization, scales as $\mathcal{O}(nm)$ per iteration and is highly parallelizable on GPUs. This parallelization enables significant speedups over the Hungarian algorithm, which has $\mathcal{O}(n^3)$ complexity, particularly in large-scale object detection tasks. RTP-DETR achieves the highest accuracy ($\sim$ 50.4 AP) while showing only minimal increases in processing time compared to efficient DINO (23 vs. 25 FPS). On average, GPU parallelization of the Sinkhorn algorithm results in a \(\sim40\%\) speedup compared to other DETR variants with Hungarian matching. These results highlight the effectiveness of optimal transport with entropy regularization for high-performance object detection.


\subsubsection{Adoption of IoU-Optimized Loss}
We explore the combination of RTP with recent DETR variants, particularly those leveraging IoU-optimized loss to enhance DETR performance significantly  \cite{liu2023detection, pu2024rank, cai2023align}. By integrating RTP with Align-DETR \cite{cai2023align}, we achieve complementary effects between RTP and IoU-aware loss. As illustrated in Table \ref{abl}, 
this combination results in an increase of \(+0.4\) AP for a 12-epoch training schedule and \(+0.7\) for a 24-epoch schedule. 

\subsubsection{Influence of Hyperparameters}
We illustrate the influence of two hyperparameters, $\epsilon_0$ and $\kappa_2$, in our matching technique. Figure \ref{abl2} details the impact of the regularization $\epsilon_0$ on the matching process. When $\epsilon_0$ is a very small value, the effect of $\epsilon$ becomes minimal, and at $\epsilon_0=0$,  entropy regularization no longer plays a role in the matching process.  Conversely, larger $\epsilon_0$ values increase the corresponding $\epsilon$, potentially risking overfitting or loss of essential details in detection. Empirical results indicate that our model performs best when $\epsilon_0$ is within the range \([0.15-0.25]\). Additionally, our analysis on the hyperparameter $\kappa_2$ is presented in Figure \ref{abl2}. Our model achieves optimal results when $\kappa_2=0.01$. We maintain $\kappa_1=1-\kappa_2$, as changing this value leads to an object imbalance problem, emphasizing its importance in preserving the accuracy of ground-truth alignment. Indeed, $\kappa_1$ ensures accurate alignment of ground truth and predictions while $\kappa_2$ allowing flexibility through probabilistic matching. If we observe too many missed ground-truth objects, we can increase $\kappa_2=0.1$. However, if the flexibility in assigning predictions is important, keeping $\kappa_2=0.01$ is fine.

\subsubsection{Visualization}
\begin{figure}
    \centering
   \includegraphics[width=1\linewidth]{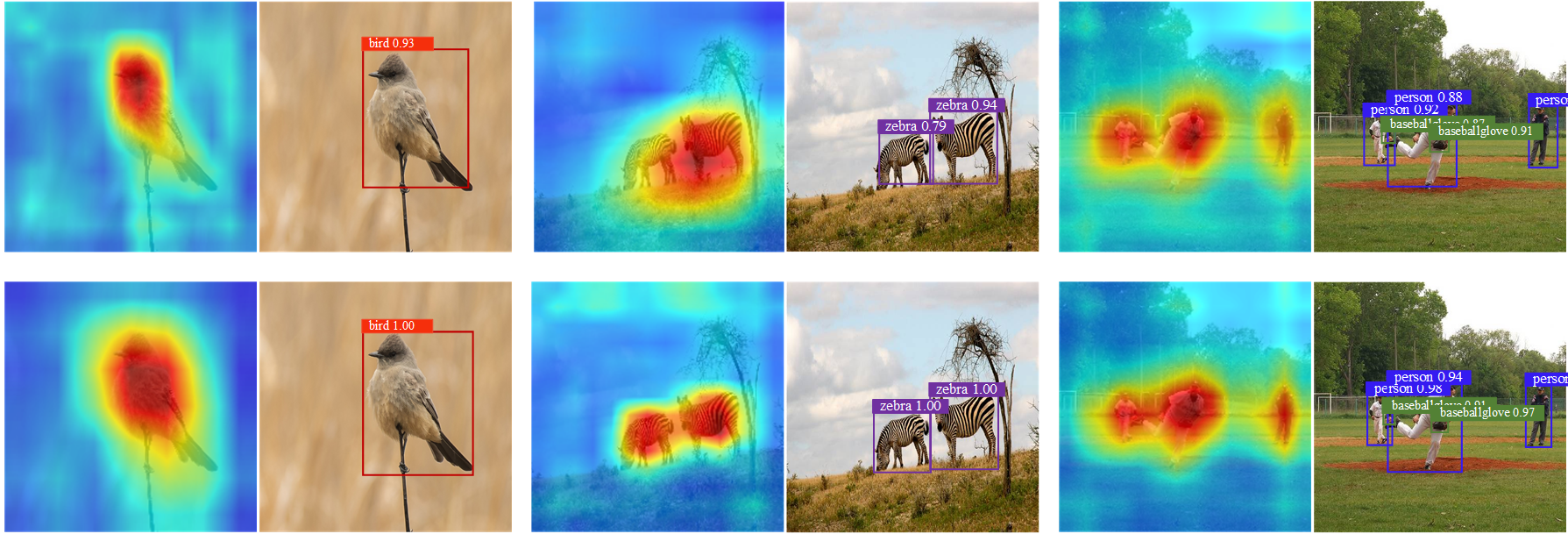}
    \caption{Visualization on sample images from COCO test set. The left side of each image displays the respective attention maps generated. The top row shows results from RTP-DETR without the regularization term, whereas the bottom row shows our full implementation. The attention maps in the second (bottom) row demonstrate that the regularized model more accurately identifies and delineates object shapes and positions, particularly in complex and densely packed scenes. }
    \label{vision}
\end{figure}

\begin{figure}
    \centering
   \includegraphics[width=1\linewidth]{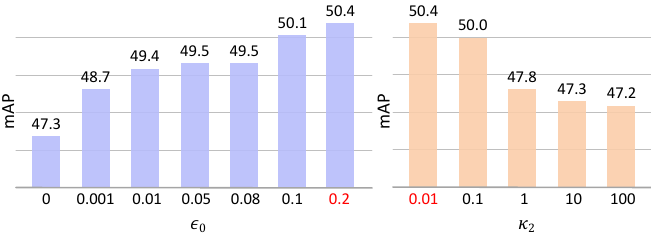}
    \caption{Influence of the hyperparameters $\epsilon$ and $\kappa_2$. The x-axis displays  $\epsilon_0$ on the left and $\kappa_2$ on the right, while the y-axis shows AP evaluated on COCO. We highlight the best-performing values for each hyperparameter in red. Importantly, if we observe too many missed ground-truth objects, we can increase $\kappa_2=0.1$. }
    \label{abl2}
\end{figure}

Two different configurations of our proposed model are visualized in Figure \ref{vision}. The corresponding attention maps for each image are presented on the left, with areas of intense red indicating the predicted object locations. The top row presents the result of the model without regularization term (RTP-DETR w/ $\epsilon=0$), representing a simpler version of the model. The bottom row displays our full implementation of the model. The distinct difference in attention map concentration between the rows is evident, particularly in the last image where the soccer field is populated with multiple small objects. The basic model struggles to handle multiple objects, resulting in scattered and unclear attention maps. In contrast, the optimized model (bottom row) exhibits well-defined and focused attention, accurately capturing the shape and position of each object, even in densely packed scenes. This demonstrates that regularization helps the model to better concentrate on relevant parts of the image, thereby improving object detection accuracy.


\section{Conclusion}
In this paper, we have examined the comparative effectiveness of the Hungarian algorithm and transportation plan in object detection, focusing on reducing the matching costs between predicted and actual data. With its strict one-to-one matching, the Hungarian algorithm is effective when the number of predictions matches the number of ground truths exactly. However, this rigid strategy limits its application and fails to capture the complexities and variations present in real-world data. In contrast, regularized optimal transport, through a probabilistic coupling, offers a flexible solution that accounts for the entire distribution of matches. This strategy not only facilitates a more comprehensive understanding of the {\it{relationships}} between predictions and ground-truths objects but also handles the discrepancies in set sizes. Our findings demonstrate that the transportation plan with entropic regularization consistently outperforms the Hungarian method by providing a more accurate and flexible alignment without relying on predefined thresholds. In future, we extend the regularized OT framework to zero-shot detection scenarios, where the model must detect objects not seen during training. This requires developing transferable transport plans that can generalize across different object categories and domains. 

\section*{Acknowledgment}
This research was supported by the Science and Technology Commission of Shanghai Municipality, China, under grant number 22DZ2229004.

\bibliographystyle{ACM-Reference-Format}
\balance
\bibliography{sample-base}


\begin{thebibliography}{44}


\ifx \showCODEN    \undefined \def \showCODEN     #1{\unskip}     \fi
\ifx \showDOI      \undefined \def \showDOI       #1{#1}\fi
\ifx \showISBNx    \undefined \def \showISBNx     #1{\unskip}     \fi
\ifx \showISBNxiii \undefined \def \showISBNxiii  #1{\unskip}     \fi
\ifx \showISSN     \undefined \def \showISSN      #1{\unskip}     \fi
\ifx \showLCCN     \undefined \def \showLCCN      #1{\unskip}     \fi
\ifx \shownote     \undefined \def \shownote      #1{#1}          \fi
\ifx \showarticletitle \undefined \def \showarticletitle #1{#1}   \fi
\ifx \showURL      \undefined \def \showURL       {\relax}        \fi
\providecommand\bibfield[2]{#2}
\providecommand\bibinfo[2]{#2}
\providecommand\natexlab[1]{#1}
\providecommand\showeprint[2][]{arXiv:#2}

\bibitem[Cai et~al\mbox{.}(2023)]%
        {cai2023align}
\bibfield{author}{\bibinfo{person}{Zhi Cai}, \bibinfo{person}{Songtao Liu}, \bibinfo{person}{Guodong Wang}, \bibinfo{person}{Zheng Ge}, \bibinfo{person}{Xiangyu Zhang}, {and} \bibinfo{person}{Di Huang}.} \bibinfo{year}{2023}\natexlab{}.
\newblock \showarticletitle{Align-DETR: Improving DETR with simple IoU-aware BCE loss}.
\newblock \bibinfo{journal}{\emph{arXiv preprint arXiv:2304.07527}} (\bibinfo{year}{2023}).
\newblock


\bibitem[Carion et~al\mbox{.}(2020)]%
        {carion2020end}
\bibfield{author}{\bibinfo{person}{Nicolas Carion}, \bibinfo{person}{Francisco Massa}, \bibinfo{person}{Gabriel Synnaeve}, \bibinfo{person}{Nicolas Usunier}, \bibinfo{person}{Alexander Kirillov}, {and} \bibinfo{person}{Sergey Zagoruyko}.} \bibinfo{year}{2020}\natexlab{}.
\newblock \showarticletitle{End-to-end object detection with transformers}. In \bibinfo{booktitle}{\emph{European conference on computer vision}}. \bibinfo{pages}{213--229}.
\newblock


\bibitem[Chang et~al\mbox{.}(2022)]%
        {chang2022unified}
\bibfield{author}{\bibinfo{person}{Wanxing Chang}, \bibinfo{person}{Ye Shi}, \bibinfo{person}{Hoang Tuan}, {and} \bibinfo{person}{Jingya Wang}.} \bibinfo{year}{2022}\natexlab{}.
\newblock \showarticletitle{Unified optimal transport framework for universal domain adaptation}.
\newblock \bibinfo{journal}{\emph{Advances in Neural Information Processing Systems}}  \bibinfo{volume}{35} (\bibinfo{year}{2022}), \bibinfo{pages}{29512--29524}.
\newblock


\bibitem[Chen et~al\mbox{.}(2022)]%
        {chen2022group}
\bibfield{author}{\bibinfo{person}{Qiang Chen}, \bibinfo{person}{Xiaokang Chen}, \bibinfo{person}{Gang Zeng}, {and} \bibinfo{person}{Jingdong Wang}.} \bibinfo{year}{2022}\natexlab{}.
\newblock \showarticletitle{Group detr: Fast training convergence with decoupled one-to-many label assignment}.
\newblock \bibinfo{journal}{\emph{arXiv preprint arXiv:2207.13085}} (\bibinfo{year}{2022}).
\newblock


\bibitem[Chuang et~al\mbox{.}(2023)]%
        {chuang2023infoot}
\bibfield{author}{\bibinfo{person}{Ching-Yao Chuang}, \bibinfo{person}{Stefanie Jegelka}, {and} \bibinfo{person}{David Alvarez-Melis}.} \bibinfo{year}{2023}\natexlab{}.
\newblock \showarticletitle{Infoot: Information maximizing optimal transport}. In \bibinfo{booktitle}{\emph{International Conference on Machine Learning}}. \bibinfo{pages}{6228--6242}.
\newblock


\bibitem[Clark et~al\mbox{.}(2022)]%
        {clark2022unified}
\bibfield{author}{\bibinfo{person}{Aidan Clark}, \bibinfo{person}{Diego de Las~Casas}, \bibinfo{person}{Aurelia Guy}, \bibinfo{person}{Arthur Mensch}, \bibinfo{person}{Michela Paganini}, \bibinfo{person}{Jordan Hoffmann}, \bibinfo{person}{Bogdan Damoc}, \bibinfo{person}{Blake Hechtman}, \bibinfo{person}{Trevor Cai}, \bibinfo{person}{Sebastian Borgeaud}, {et~al\mbox{.}}} \bibinfo{year}{2022}\natexlab{}.
\newblock \showarticletitle{Unified scaling laws for routed language models}. In \bibinfo{booktitle}{\emph{International conference on machine learning}}. PMLR, \bibinfo{pages}{4057--4086}.
\newblock


\bibitem[Cuturi(2013)]%
        {cuturi2013sinkhorn}
\bibfield{author}{\bibinfo{person}{Marco Cuturi}.} \bibinfo{year}{2013}\natexlab{}.
\newblock \showarticletitle{Sinkhorn distances: Lightspeed computation of optimal transport}.
\newblock \bibinfo{journal}{\emph{Advances in neural information processing systems}}  \bibinfo{volume}{26} (\bibinfo{year}{2013}).
\newblock


\bibitem[De~Plaen et~al\mbox{.}(2023)]%
        {De_Plaen_2023_CVPR}
\bibfield{author}{\bibinfo{person}{Henri De~Plaen}, \bibinfo{person}{Pierre-Fran\c{c}ois De~Plaen}, \bibinfo{person}{Johan A.~K. Suykens}, \bibinfo{person}{Marc Proesmans}, \bibinfo{person}{Tinne Tuytelaars}, {and} \bibinfo{person}{Luc Van~Gool}.} \bibinfo{year}{2023}\natexlab{}.
\newblock \showarticletitle{Unbalanced Optimal Transport: A Unified Framework for Object Detection}. In \bibinfo{booktitle}{\emph{Proceedings of the IEEE/CVF Conference on Computer Vision and Pattern Recognition}}. \bibinfo{pages}{3198--3207}.
\newblock


\bibitem[Detector(2022)]%
        {detector2022fcos}
\bibfield{author}{\bibinfo{person}{Anchor-Free~Object Detector}.} \bibinfo{year}{2022}\natexlab{}.
\newblock \showarticletitle{FCOS: A Simple and Strong Anchor-Free Object Detector}.
\newblock \bibinfo{journal}{\emph{IEEE Transactions on Pattern Analysis and Machine Intelligence}} \bibinfo{volume}{44}, \bibinfo{number}{4} (\bibinfo{year}{2022}).
\newblock


\bibitem[Gao et~al\mbox{.}(2021)]%
        {gao2021fast}
\bibfield{author}{\bibinfo{person}{Peng Gao}, \bibinfo{person}{Minghang Zheng}, \bibinfo{person}{Xiaogang Wang}, \bibinfo{person}{Jifeng Dai}, {and} \bibinfo{person}{Hongsheng Li}.} \bibinfo{year}{2021}\natexlab{}.
\newblock \showarticletitle{Fast convergence of detr with spatially modulated co-attention}. In \bibinfo{booktitle}{\emph{Proceedings of the IEEE/CVF international conference on computer vision}}. \bibinfo{pages}{3621--3630}.
\newblock


\bibitem[Ge et~al\mbox{.}(2021)]%
        {ge2021ota}
\bibfield{author}{\bibinfo{person}{Zheng Ge}, \bibinfo{person}{Songtao Liu}, \bibinfo{person}{Zeming Li}, \bibinfo{person}{Osamu Yoshie}, {and} \bibinfo{person}{Jian Sun}.} \bibinfo{year}{2021}\natexlab{}.
\newblock \showarticletitle{Ota: Optimal transport assignment for object detection}. In \bibinfo{booktitle}{\emph{Proceedings of the IEEE/CVF conference on computer vision and pattern recognition}}. \bibinfo{pages}{303--312}.
\newblock


\bibitem[He et~al\mbox{.}(2016)]%
        {he2016deep}
\bibfield{author}{\bibinfo{person}{Kaiming He}, \bibinfo{person}{Xiangyu Zhang}, \bibinfo{person}{Shaoqing Ren}, {and} \bibinfo{person}{Jian Sun}.} \bibinfo{year}{2016}\natexlab{}.
\newblock \showarticletitle{Deep residual learning for image recognition}. In \bibinfo{booktitle}{\emph{Proceedings of the IEEE conference on computer vision and pattern recognition}}. \bibinfo{pages}{770--778}.
\newblock


\bibitem[Hou et~al\mbox{.}(2024)]%
        {hou2024salience}
\bibfield{author}{\bibinfo{person}{Xiuquan Hou}, \bibinfo{person}{Meiqin Liu}, \bibinfo{person}{Senlin Zhang}, \bibinfo{person}{Ping Wei}, {and} \bibinfo{person}{Badong Chen}.} \bibinfo{year}{2024}\natexlab{}.
\newblock \showarticletitle{Salience DETR: Enhancing Detection Transformer with Hierarchical Salience Filtering Refinement}.
\newblock \bibinfo{journal}{\emph{Proceedings of the IEEE/CVF conference on computer vision and pattern recognition}} (\bibinfo{year}{2024}).
\newblock


\bibitem[Jia et~al\mbox{.}(2023)]%
        {jia2023detrs}
\bibfield{author}{\bibinfo{person}{Ding Jia}, \bibinfo{person}{Yuhui Yuan}, \bibinfo{person}{Haodi He}, \bibinfo{person}{Xiaopei Wu}, \bibinfo{person}{Haojun Yu}, \bibinfo{person}{Weihong Lin}, \bibinfo{person}{Lei Sun}, \bibinfo{person}{Chao Zhang}, {and} \bibinfo{person}{Han Hu}.} \bibinfo{year}{2023}\natexlab{}.
\newblock \showarticletitle{Detrs with hybrid matching}. In \bibinfo{booktitle}{\emph{Proceedings of the IEEE/CVF conference on computer vision and pattern recognition}}. \bibinfo{pages}{19702--19712}.
\newblock


\bibitem[Kuhn(1955)]%
        {kuhn1955hungarian}
\bibfield{author}{\bibinfo{person}{Harold~W Kuhn}.} \bibinfo{year}{1955}\natexlab{}.
\newblock \showarticletitle{The Hungarian method for the assignment problem}.
\newblock \bibinfo{journal}{\emph{Naval research logistics quarterly}} \bibinfo{volume}{2}, \bibinfo{number}{1-2} (\bibinfo{year}{1955}), \bibinfo{pages}{83--97}.
\newblock


\bibitem[Li et~al\mbox{.}(2022)]%
        {li2022dn}
\bibfield{author}{\bibinfo{person}{Feng Li}, \bibinfo{person}{Hao Zhang}, \bibinfo{person}{Shilong Liu}, \bibinfo{person}{Jian Guo}, \bibinfo{person}{Lionel~M Ni}, {and} \bibinfo{person}{Lei Zhang}.} \bibinfo{year}{2022}\natexlab{}.
\newblock \showarticletitle{Dn-detr: Accelerate detr training by introducing query denoising}. In \bibinfo{booktitle}{\emph{Proceedings of the IEEE/CVF conference on computer vision and pattern recognition}}. \bibinfo{pages}{13619--13627}.
\newblock


\bibitem[Lin et~al\mbox{.}(2022)]%
        {lin2022learning}
\bibfield{author}{\bibinfo{person}{Chuang Lin}, \bibinfo{person}{Peize Sun}, \bibinfo{person}{Yi Jiang}, \bibinfo{person}{Ping Luo}, \bibinfo{person}{Lizhen Qu}, \bibinfo{person}{Gholamreza Haffari}, \bibinfo{person}{Zehuan Yuan}, {and} \bibinfo{person}{Jianfei Cai}.} \bibinfo{year}{2022}\natexlab{}.
\newblock \showarticletitle{Learning object-language alignments for open-vocabulary object detection}.
\newblock \bibinfo{journal}{\emph{arXiv preprint arXiv:2211.14843}} (\bibinfo{year}{2022}).
\newblock


\bibitem[Lin et~al\mbox{.}(2014)]%
        {lin2014microsoft}
\bibfield{author}{\bibinfo{person}{Tsung-Yi Lin}, \bibinfo{person}{Michael Maire}, \bibinfo{person}{Serge Belongie}, \bibinfo{person}{James Hays}, \bibinfo{person}{Pietro Perona}, \bibinfo{person}{Deva Ramanan}, \bibinfo{person}{Piotr Doll{\'a}r}, {and} \bibinfo{person}{C~Lawrence Zitnick}.} \bibinfo{year}{2014}\natexlab{}.
\newblock \showarticletitle{Microsoft coco: Common objects in context}. In \bibinfo{booktitle}{\emph{European Conference on Computer Vision}}. \bibinfo{pages}{740--755}.
\newblock


\bibitem[Liu et~al\mbox{.}(2023c)]%
        {liu2023zero}
\bibfield{author}{\bibinfo{person}{Huan Liu}, \bibinfo{person}{Lu Zhang}, \bibinfo{person}{Jihong Guan}, {and} \bibinfo{person}{Shuigeng Zhou}.} \bibinfo{year}{2023}\natexlab{c}.
\newblock \showarticletitle{Zero-Shot Object Detection by Semantics-Aware DETR with Adaptive Contrastive Loss}. In \bibinfo{booktitle}{\emph{Proceedings of the ACM International Conference on Multimedia}}. \bibinfo{pages}{4421--4430}.
\newblock


\bibitem[Liu et~al\mbox{.}(2023b)]%
        {liu2023detection}
\bibfield{author}{\bibinfo{person}{Shilong Liu}, \bibinfo{person}{Tianhe Ren}, \bibinfo{person}{Jiayu Chen}, \bibinfo{person}{Zhaoyang Zeng}, \bibinfo{person}{Hao Zhang}, \bibinfo{person}{Feng Li}, \bibinfo{person}{Hongyang Li}, \bibinfo{person}{Jun Huang}, \bibinfo{person}{Hang Su}, {and} \bibinfo{person}{Jun Zhu}.} \bibinfo{year}{2023}\natexlab{b}.
\newblock \showarticletitle{Detection transformer with stable matching}. In \bibinfo{booktitle}{\emph{Proceedings of the IEEE/CVF International Conference on Computer Vision}}. \bibinfo{pages}{6491--6500}.
\newblock


\bibitem[Liu et~al\mbox{.}(2023a)]%
        {liu2022sparsity}
\bibfield{author}{\bibinfo{person}{Tianlin Liu}, \bibinfo{person}{Joan Puigcerver}, {and} \bibinfo{person}{Mathieu Blondel}.} \bibinfo{year}{2023}\natexlab{a}.
\newblock \showarticletitle{Sparsity-constrained optimal transport}.
\newblock \bibinfo{journal}{\emph{International Conference on Learning Representations}} (\bibinfo{year}{2023}).
\newblock


\bibitem[Liu et~al\mbox{.}(2016)]%
        {liu2016ssd}
\bibfield{author}{\bibinfo{person}{Wei Liu}, \bibinfo{person}{Dragomir Anguelov}, \bibinfo{person}{Dumitru Erhan}, \bibinfo{person}{Christian Szegedy}, \bibinfo{person}{Scott Reed}, \bibinfo{person}{Cheng-Yang Fu}, {and} \bibinfo{person}{Alexander~C Berg}.} \bibinfo{year}{2016}\natexlab{}.
\newblock \showarticletitle{Ssd: Single shot multibox detector}. In \bibinfo{booktitle}{\emph{European Conference on Computer Vision}}. \bibinfo{pages}{21--37}.
\newblock


\bibitem[Liu et~al\mbox{.}(2021)]%
        {liu2021swin}
\bibfield{author}{\bibinfo{person}{Ze Liu}, \bibinfo{person}{Yutong Lin}, \bibinfo{person}{Yue Cao}, \bibinfo{person}{Han Hu}, \bibinfo{person}{Yixuan Wei}, \bibinfo{person}{Zheng Zhang}, \bibinfo{person}{Stephen Lin}, {and} \bibinfo{person}{Baining Guo}.} \bibinfo{year}{2021}\natexlab{}.
\newblock \showarticletitle{Swin transformer: Hierarchical vision transformer using shifted windows}. In \bibinfo{booktitle}{\emph{Proceedings of the IEEE/CVF international conference on computer vision}}. \bibinfo{pages}{10012--10022}.
\newblock


\bibitem[Paty and Cuturi(2020)]%
        {paty2020regularized}
\bibfield{author}{\bibinfo{person}{Fran{\c{c}}ois-Pierre Paty} {and} \bibinfo{person}{Marco Cuturi}.} \bibinfo{year}{2020}\natexlab{}.
\newblock \showarticletitle{Regularized optimal transport is ground cost adversarial}. In \bibinfo{booktitle}{\emph{International Conference on Machine Learning}}. PMLR, \bibinfo{pages}{7532--7542}.
\newblock


\bibitem[Peyr{\'e} et~al\mbox{.}(2019)]%
        {peyre2019computational}
\bibfield{author}{\bibinfo{person}{Gabriel Peyr{\'e}}, \bibinfo{person}{Marco Cuturi}, {et~al\mbox{.}}} \bibinfo{year}{2019}\natexlab{}.
\newblock \showarticletitle{Computational optimal transport: With applications to data science}.
\newblock \bibinfo{journal}{\emph{Foundations and Trends{\textregistered} in Machine Learning}} \bibinfo{volume}{11}, \bibinfo{number}{5-6} (\bibinfo{year}{2019}), \bibinfo{pages}{355--607}.
\newblock


\bibitem[Pu et~al\mbox{.}(2024)]%
        {pu2024rank}
\bibfield{author}{\bibinfo{person}{Yifan Pu}, \bibinfo{person}{Weicong Liang}, \bibinfo{person}{Yiduo Hao}, \bibinfo{person}{Yuhui Yuan}, \bibinfo{person}{Yukang Yang}, \bibinfo{person}{Chao Zhang}, \bibinfo{person}{Han Hu}, {and} \bibinfo{person}{Gao Huang}.} \bibinfo{year}{2024}\natexlab{}.
\newblock \showarticletitle{Rank-DETR for high quality object detection}.
\newblock \bibinfo{journal}{\emph{Advances in Neural Information Processing Systems}}  \bibinfo{volume}{36} (\bibinfo{year}{2024}).
\newblock


\bibitem[Redmon et~al\mbox{.}(2016)]%
        {redmon2016you}
\bibfield{author}{\bibinfo{person}{Joseph Redmon}, \bibinfo{person}{Santosh Divvala}, \bibinfo{person}{Ross Girshick}, {and} \bibinfo{person}{Ali Farhadi}.} \bibinfo{year}{2016}\natexlab{}.
\newblock \showarticletitle{You only look once: Unified, real-time object detection}. In \bibinfo{booktitle}{\emph{Proceedings of the IEEE conference on computer vision and pattern recognition}}. \bibinfo{pages}{779--788}.
\newblock


\bibitem[Ren et~al\mbox{.}(2015)]%
        {ren2015faster}
\bibfield{author}{\bibinfo{person}{Shaoqing Ren}, \bibinfo{person}{Kaiming He}, \bibinfo{person}{Ross Girshick}, {and} \bibinfo{person}{Jian Sun}.} \bibinfo{year}{2015}\natexlab{}.
\newblock \showarticletitle{Faster r-cnn: Towards real-time object detection with region proposal networks}.
\newblock \bibinfo{journal}{\emph{Advances in neural information processing systems}}  \bibinfo{volume}{28} (\bibinfo{year}{2015}).
\newblock


\bibitem[Roh et~al\mbox{.}(2021)]%
        {roh2021sparse}
\bibfield{author}{\bibinfo{person}{Byungseok Roh}, \bibinfo{person}{JaeWoong Shin}, \bibinfo{person}{Wuhyun Shin}, {and} \bibinfo{person}{Saehoon Kim}.} \bibinfo{year}{2021}\natexlab{}.
\newblock \showarticletitle{Sparse detr: Efficient end-to-end object detection with learnable sparsity}.
\newblock \bibinfo{journal}{\emph{arXiv preprint arXiv:2111.14330}} (\bibinfo{year}{2021}).
\newblock


\bibitem[Sebbouh et~al\mbox{.}(2023)]%
        {sebbouh2023structured}
\bibfield{author}{\bibinfo{person}{Othmane Sebbouh}, \bibinfo{person}{Marco Cuturi}, {and} \bibinfo{person}{Gabriel Peyr{\'e}}.} \bibinfo{year}{2023}\natexlab{}.
\newblock \showarticletitle{Structured Transforms Across Spaces with Cost-Regularized Optimal Transport}.
\newblock \bibinfo{journal}{\emph{arXiv preprint arXiv:2311.05788}} (\bibinfo{year}{2023}).
\newblock


\bibitem[Shamsolmoali et~al\mbox{.}(2022)]%
        {shamsolmoali2022gen}
\bibfield{author}{\bibinfo{person}{Pourya Shamsolmoali}, \bibinfo{person}{Masoumeh Zareapoor}, \bibinfo{person}{Swagatam Das}, \bibinfo{person}{Salvador Garcia}, \bibinfo{person}{Eric Granger}, {and} \bibinfo{person}{Jie Yang}.} \bibinfo{year}{2022}\natexlab{}.
\newblock \showarticletitle{GEN: Generative equivariant networks for diverse image-to-image translation}.
\newblock \bibinfo{journal}{\emph{IEEE Transactions on Cybernetics}} \bibinfo{volume}{53}, \bibinfo{number}{2} (\bibinfo{year}{2022}), \bibinfo{pages}{874--886}.
\newblock


\bibitem[Shamsolmoali et~al\mbox{.}(2024a)]%
        {shamsolmoali2024hybrid}
\bibfield{author}{\bibinfo{person}{Pourya Shamsolmoali}, \bibinfo{person}{Masoumeh Zareapoor}, \bibinfo{person}{Swagatam Das}, \bibinfo{person}{Eric Granger}, {and} \bibinfo{person}{Salvador Garcia}.} \bibinfo{year}{2024}\natexlab{a}.
\newblock \showarticletitle{Hybrid Gromov--Wasserstein Embedding for Capsule Learning}.
\newblock \bibinfo{journal}{\emph{IEEE Transactions on Neural Networks and Learning Systems}} (\bibinfo{year}{2024}).
\newblock


\bibitem[Shamsolmoali et~al\mbox{.}(2024b)]%
        {shamsolmoali2024setformer}
\bibfield{author}{\bibinfo{person}{Pourya Shamsolmoali}, \bibinfo{person}{Masoumeh Zareapoor}, \bibinfo{person}{Eric Granger}, {and} \bibinfo{person}{Michael Felsberg}.} \bibinfo{year}{2024}\natexlab{b}.
\newblock \showarticletitle{SeTformer is what you need for vision and language}. In \bibinfo{booktitle}{\emph{Proceedings of the AAAI Conference on Artificial Intelligence}}, Vol.~\bibinfo{volume}{38}. \bibinfo{pages}{4713--4721}.
\newblock


\bibitem[Sun et~al\mbox{.}(2021)]%
        {sun2021sparse}
\bibfield{author}{\bibinfo{person}{Peize Sun}, \bibinfo{person}{Rufeng Zhang}, \bibinfo{person}{Yi Jiang}, \bibinfo{person}{Tao Kong}, \bibinfo{person}{Chenfeng Xu}, \bibinfo{person}{Wei Zhan}, \bibinfo{person}{Masayoshi Tomizuka}, \bibinfo{person}{Lei Li}, \bibinfo{person}{Zehuan Yuan}, \bibinfo{person}{Changhu Wang}, {et~al\mbox{.}}} \bibinfo{year}{2021}\natexlab{}.
\newblock \showarticletitle{Sparse r-cnn: End-to-end object detection with learnable proposals}. In \bibinfo{booktitle}{\emph{Proceedings of the IEEE/CVF conference on computer vision and pattern recognition}}. \bibinfo{pages}{14454--14463}.
\newblock


\bibitem[Wei et~al\mbox{.}(2023)]%
        {wei2023guide}
\bibfield{author}{\bibinfo{person}{Kaiwen Wei}, \bibinfo{person}{Yiran Yang}, \bibinfo{person}{Li Jin}, \bibinfo{person}{Xian Sun}, \bibinfo{person}{Zequn Zhang}, \bibinfo{person}{Jingyuan Zhang}, \bibinfo{person}{Xiao Li}, \bibinfo{person}{Linhao Zhang}, \bibinfo{person}{Jintao Liu}, {and} \bibinfo{person}{Guo Zhi}.} \bibinfo{year}{2023}\natexlab{}.
\newblock \showarticletitle{Guide the many-to-one assignment: Open information extraction via iou-aware optimal transport}. In \bibinfo{booktitle}{\emph{Proceedings of the 61st Annual Meeting of the Association for Computational Linguistics (Volume 1: Long Papers)}}. \bibinfo{pages}{4971--4984}.
\newblock


\bibitem[Wong et~al\mbox{.}(2019)]%
        {wong2019wasserstein}
\bibfield{author}{\bibinfo{person}{Eric Wong}, \bibinfo{person}{Frank Schmidt}, {and} \bibinfo{person}{Zico Kolter}.} \bibinfo{year}{2019}\natexlab{}.
\newblock \showarticletitle{Wasserstein adversarial examples via projected sinkhorn iterations}. In \bibinfo{booktitle}{\emph{International conference on machine learning}}. \bibinfo{pages}{6808--6817}.
\newblock


\bibitem[Xie et~al\mbox{.}(2022)]%
        {xie2022solving}
\bibfield{author}{\bibinfo{person}{Yiling Xie}, \bibinfo{person}{Yiling Luo}, {and} \bibinfo{person}{Xiaoming Huo}.} \bibinfo{year}{2022}\natexlab{}.
\newblock \showarticletitle{Solving a special type of optimal transport problem by a modified Hungarian algorithm}.
\newblock \bibinfo{journal}{\emph{arXiv preprint arXiv:2210.16645}} (\bibinfo{year}{2022}).
\newblock


\bibitem[Yao et~al\mbox{.}(2021)]%
        {yao2021efficient}
\bibfield{author}{\bibinfo{person}{Zhuyu Yao}, \bibinfo{person}{Jiangbo Ai}, \bibinfo{person}{Boxun Li}, {and} \bibinfo{person}{Chi Zhang}.} \bibinfo{year}{2021}\natexlab{}.
\newblock \showarticletitle{Efficient detr: improving end-to-end object detector with dense prior}.
\newblock \bibinfo{journal}{\emph{arXiv preprint arXiv:2104.01318}} (\bibinfo{year}{2021}).
\newblock


\bibitem[Zhang et~al\mbox{.}(2022a)]%
        {zhang2022accelerating}
\bibfield{author}{\bibinfo{person}{Gongjie Zhang}, \bibinfo{person}{Zhipeng Luo}, \bibinfo{person}{Yingchen Yu}, \bibinfo{person}{Kaiwen Cui}, {and} \bibinfo{person}{Shijian Lu}.} \bibinfo{year}{2022}\natexlab{a}.
\newblock \showarticletitle{Accelerating DETR convergence via semantic-aligned matching}. In \bibinfo{booktitle}{\emph{Proceedings of the IEEE/CVF conference on computer vision and pattern recognition}}. \bibinfo{pages}{949--958}.
\newblock


\bibitem[Zhang et~al\mbox{.}(2023)]%
        {zhang2022dino}
\bibfield{author}{\bibinfo{person}{Hao Zhang}, \bibinfo{person}{Feng Li}, \bibinfo{person}{Shilong Liu}, \bibinfo{person}{Lei Zhang}, \bibinfo{person}{Hang Su}, \bibinfo{person}{Jun Zhu}, \bibinfo{person}{Lionel~M Ni}, {and} \bibinfo{person}{Heung-Yeung Shum}.} \bibinfo{year}{2023}\natexlab{}.
\newblock \showarticletitle{Dino: Detr with improved denoising anchor boxes for end-to-end object detection}.
\newblock \bibinfo{journal}{\emph{The International Conference on Learning Representations}} (\bibinfo{year}{2023}).
\newblock


\bibitem[Zhang et~al\mbox{.}(2022b)]%
        {zhang2022robust}
\bibfield{author}{\bibinfo{person}{Yifu Zhang}, \bibinfo{person}{Chunyu Wang}, \bibinfo{person}{Xinggang Wang}, \bibinfo{person}{Wenjun Zeng}, {and} \bibinfo{person}{Wenyu Liu}.} \bibinfo{year}{2022}\natexlab{b}.
\newblock \showarticletitle{Robust multi-object tracking by marginal inference}. In \bibinfo{booktitle}{\emph{European Conference on Computer Vision}}. \bibinfo{pages}{22--40}.
\newblock


\bibitem[Zhao et~al\mbox{.}(2024)]%
        {zhao2024ms}
\bibfield{author}{\bibinfo{person}{Chuyang Zhao}, \bibinfo{person}{Yifan Sun}, \bibinfo{person}{Wenhao Wang}, \bibinfo{person}{Qiang Chen}, \bibinfo{person}{Errui Ding}, \bibinfo{person}{Yi Yang}, {and} \bibinfo{person}{Jingdong Wang}.} \bibinfo{year}{2024}\natexlab{}.
\newblock \showarticletitle{MS-DETR: Efficient DETR Training with Mixed Supervision}.
\newblock \bibinfo{journal}{\emph{Proceedings of the IEEE/CVF conference on computer vision and pattern recognition}} (\bibinfo{year}{2024}).
\newblock


\bibitem[Zhao et~al\mbox{.}(2019)]%
        {zhao2019object}
\bibfield{author}{\bibinfo{person}{Zhong-Qiu Zhao}, \bibinfo{person}{Peng Zheng}, \bibinfo{person}{Shou-tao Xu}, {and} \bibinfo{person}{Xindong Wu}.} \bibinfo{year}{2019}\natexlab{}.
\newblock \showarticletitle{Object detection with deep learning: A review}.
\newblock \bibinfo{journal}{\emph{IEEE transactions on neural networks and learning systems}} \bibinfo{volume}{30}, \bibinfo{number}{11} (\bibinfo{year}{2019}), \bibinfo{pages}{3212--3232}.
\newblock


\bibitem[Zhu et~al\mbox{.}(2020)]%
        {zhu2020deformable}
\bibfield{author}{\bibinfo{person}{Xizhou Zhu}, \bibinfo{person}{Weijie Su}, \bibinfo{person}{Lewei Lu}, \bibinfo{person}{Bin Li}, \bibinfo{person}{Xiaogang Wang}, {and} \bibinfo{person}{Jifeng Dai}.} \bibinfo{year}{2020}\natexlab{}.
\newblock \showarticletitle{Deformable detr: Deformable transformers for end-to-end object detection}.
\newblock \bibinfo{journal}{\emph{The International Conference on Learning Representations}} (\bibinfo{year}{2020}).
\newblock


\end{thebibliography}

\end{document}